\documentclass[sigconf]{acmart}
\usepackage{array}
\usepackage{amsmath}
\usepackage{amsthm}
\usepackage{subfig}
\usepackage{enumitem}
\usepackage{framed}
\usepackage{graphicx}
\usepackage{color}
\usepackage{multirow}
\usepackage{algorithm}
\usepackage{algpseudocode}
\usepackage{pgfplots}
\usepackage{pgfplotstable}
\pgfplotsset{compat=1.18}
\usepackage{adjustbox}
\newtheorem{definition}{Definition} 

\begin{document}

\title{Learning Individual Movement Shifts After Urban Disruptions with Social Infrastructure Reliance}

\author{Shangde Gao}
\orcid{0000-0003-2218-2872}
\authornote{Corresponding author.}
\authornote{Both authors contributed equally to this research.}
\email{gao.shangde@ufl.edu}
\affiliation{%
  \institution{School of Architecture}
  \institution{University of Florida}
  \city{Gainesville}
  \state{Florida}
  \country{USA}
}

\author{Zelin Xu}
\orcid{0009-0004-4419-3155}
\authornotemark[1]
\authornotemark[2]
\email{zelin.xu@ufl.edu}
\affiliation{%
  \institution{Department of CISE}
  \institution{University of Florida}
  \city{Gainesville}
  \state{Florida}
  \country{USA}
}

\author{Zhe Jiang}
\orcid{0000-0002-3576-6976}
\email{zhe.jiang@ufl.edu}
\affiliation{%
  \institution{Department of CISE}
  \institution{University of Florida}
  \city{Gainesville}
  \state{Florida}
  \country{USA}
}

\begin{abstract}
    Shifts in individual movement patterns following disruptive events can reveal changing demands for community resources. 
However, predicting such shifts before disruptive events remains challenging for several reasons. 
First, measures are lacking for individuals’ heterogeneous social infrastructure resilience (SIR), which directly influences their movement patterns, and commonly used features are often limited or unavailable at scale, e.g., sociodemographic characteristics. 
Second, 
the complex interactions between individual movement patterns and spatial contexts have not been sufficiently captured. 
Third, 
individual-level movement may be 
spatially sparse and not well-suited to traditional decision-making methods for movement predictions.
This study incorporates individuals’ SIR into a conditioned deep learning model to capture the complex relationships between individual movement patterns and local spatial context using large-scale, sparse individual-level data.
Our experiments 
demonstrate that incorporating individuals' SIR and spatial context can enhance the model's ability to predict post-event individual movement patterns.
The conditioned model can capture the divergent shifts in movement patterns among individuals who exhibit similar pre-event patterns but differ in SIR.
\end{abstract}



\maketitle

\section{Introduction}
Urban dynamics are fundamentally shaped by human movements, which encapsulate how people access daily-life resources \cite{gao_extracting_2017}. 
Understanding and predicting human movement patterns not only reveal the spatial and temporal rhythms of local communities but also support the development of sustainable and resilient urban environments \cite{haraguchi_human_2022}.
This becomes critical when communities confront disruptive events---such as natural disasters, pandemics, or infrastructure failures---which may abruptly affect human movements and stress urban systems~\cite{gong_uncovering_2025}. 
Predicting movement pattern shifts after disruptive events can help prepare for future event responses by anticipating population needs, guiding emergency response, and strengthening urban adaptability in the post-event normal \cite{hong_measuring_2021}. 

\begin{figure}
    \centering
    \includegraphics[width=0.45\textwidth]{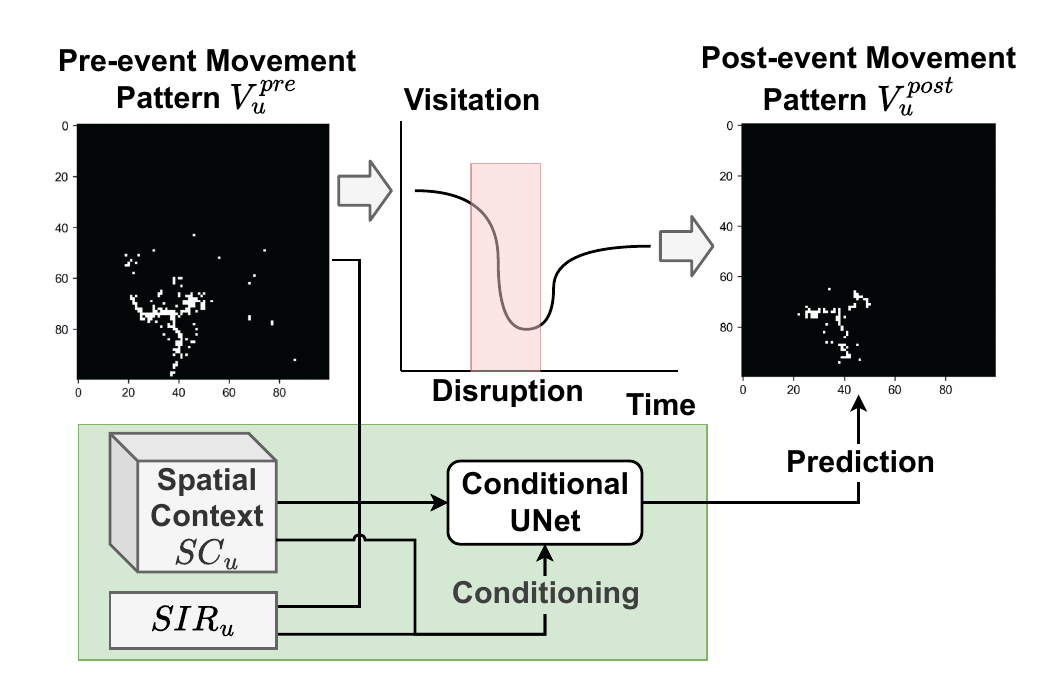}
    \caption{Illustration of individual movement pattern shifts after urban disruptive events.}
    \label{fig:mobilitychange}
\end{figure}

Previous studies have focused on population-level movement patterns after natural disasters and pandemics  \cite{lara_modeling_2023,gong_uncovering_2025}.
While informative for community-level response and adaptation \cite{haraguchi_human_2022}, such an approach may obscure individual-level heterogeneity \cite{yang_identifying_2023}. 
As Figure~\ref{fig:mobilitychange} shows, some people may drastically alter movement patterns after disruptions, while others may adjust minimally \cite{zhang_characterizing_2022}. Such heterogeneity can be related to factors of individuals, such as place of residence and surrounding built environment \cite{gao_data-driven_2023}, preference in daily activities \cite{zhang_characterizing_2022}, and social infrastructure reliance (SIR in short) \cite{li_service_2024}. 
Accounting for this individual-level heterogeneity can complement population-level analyses and improve urban resilience planning by revealing nuanced behavioral responses, identifying vulnerable groups, and guiding targeted interventions. For example, individuals with poor health conditions may increase visits to health-related social infrastructure, yet these movement shifts can be masked within overall population patterns.

Despite its importance, individuals' movement patterns and shifts after disruptive events remain underexplored due to several key challenges. 
First, widely-used traits to predict individual movement patterns, such as socio-demographic attributes, may not predict mobility patterns accurately \cite{yang_identifying_2023} and are difficult to obtain at scale due to privacy and cost constraints \cite{engin_data-driven_2020}. 
In contrast, SIR inferred from historical movement patterns \cite{zhao_unravelling_2024} can better signal shifts: high-SIR individuals may maintain stable movement patterns, while low-SIR individuals can change movements dynamically \cite{zhang_characterizing_2022}. 
Yet, SIR has been largely overlooked in current models of post-disruption movement patterns.
Second, individual movement patterns can be highly context-dependent \cite{gong_uncovering_2025}, shaped by the spatial context of the local built environment, such as access to healthcare, transit, and other services \cite{gao_data-driven_2023}. 
Most existing models may not capture the complex interactions between individual movement patterns and spatial context, or they lack the architectural flexibility to integrate spatial context into individual movement pattern predictions. 
Third, data of individual movement patterns is often sparse in both spatial and temporal perspectives \cite{yabe_mobile_2022}, which may not be well-suited to traditional decision-making methods for movement predictions. 

This study aims to predict the shifts in individual movement patterns before and after disruptive events, conditioned on individuals' SIR and the spatial context. 
SIR is measured as an individual's likelihood of increasing or decreasing visits to certain social infrastructure when confronting and responding to the disruptions.
To address this problem, we employ a conditional UNet model \cite{esser_variational_2018}, referred to as cUNet. 
The cUNet model learns the shifts of individuals' movement patterns from normal periods before disruptive events to the post-event patterns, conditioned on individuals' SIR across types of social infrastructure and the spatial context of their place of residence. We evaluate the cUNet model using YJMob100K \cite{yabe_yjmob100k_2024}, a large-scale real-world mobility dataset enriched with detailed social infrastructure distribution, aiming to demonstrate the model's ability to capture nuanced shifts in individual movement patterns and offer a foundation for future advancements in behavior-aware and context-adaptive movement models.

In summary, the contributions of this paper are as follows:
\begin{itemize}[noitemsep, topsep=0pt, leftmargin=*]
\item \textbf{Integrating SIR and spatial context into predictions of shifts of individual movement patterns:} We incorporate SIR and fine-grained spatial contexts as conditioning inputs, enabling our model to predict nuanced changes in individual-level movement patterns after disruptive events, including divergent behavioral responses among individuals with similar pre-event patterns but differing SIR. This integration can enhance the prediction’s accuracy and interpretability for future behavior-aware urban resilience modeling.
\item \textbf{Capturing the complex interactions between individual movement patterns and spatial context:} We employ conditioned deep learning models to learn the influence of individual SIR and local spatial context on post-event movement patterns. This approach enables the model to represent heterogeneous behavioral responses conditioned on individual-level and localized traits, which are usually difficult to capture using traditional or population-level models.
\item \textbf{Employing a conditional deep learning model to learn from sparse data of individual movement patterns:} Our UNet-based architecture with spatial and SIR conditioning can effectively learn individual-level movement pattern changes from limited and sparse individual-level data, enabling robust generalization across diverse populations.
\end{itemize}

\section{Problem Statement}\label{sec:problem}

\begin{definition}
\textbf{Individual Movement Pattern} is the spatial footprint of an individual’s activities within a region over a given time period. It captures where an individual has physically visited and serves as a proxy for behavioral routines, spatial preferences, and interactions with the built environment. For an individual $u$, the pattern is represented by a binary spatial grid $\mathbf{V}_u \in \{0,1\}^{M \times N}$, where $v_{u,i,j} = 1$ if $u$ visited cell $(i,j)$ during the observation period, and $v_{u,i,j} = 0$ otherwise. Let $\mathcal{U}$ denote the set of individuals, each associated with a movement pattern $\mathbf{V}_u$.
\end{definition}

\begin{definition}
\textbf{Social Infrastructure Reliance (SIR)} captures an individual’s dependence on social infrastructure (e.g., grocery stores, healthcare, schools, recreation) and potential sensitivity to their availability. It reflects how an individual's movement patterns may shift in response to disruptions affecting these services. For each individual $u$, we define a vector $\mathbf{SIR}_u \in \mathbb{R}^K$, where $K$ is the number of infrastructure categories and the $k$-th entry reflects $u$’s reliance on infrastructure type $k$. In this work, $\mathbf{SIR}_u$ is estimated from the frequency of historical visits to Points of Interest (POIs), providing a compact behavioral profile that conditions prediction of post-disruption movement patterns.
\end{definition}

\begin{definition}
\textbf{Spatial Context} describes the built environment around an individual, capturing the distribution and composition of nearby services. It reflects structural features such as the diversity and density of land use and serves as a key factor influencing individual movement patterns, especially under disruptive events. For each individual $u$, we define $\mathbf{SC}_u \in \mathbb{R}^{K \times G \times G}$, where $K$ is the number of infrastructure categories and $G \times G$ is a grid centered at $u$’s home. Each cell $(i,j)$ stores a vector $\mathbf{SC}_{u,i,j} \in \mathbb{R}^K$, with the $k$-th entry denoting the proportion of POIs of type $k$ within that cell. This representation summarizes local service heterogeneity for modeling movement responses to disruptions.
\end{definition}



\textbf{Problem Formulation.}  
Given an individual $u \in \mathcal{U}$ with pre-event movement pattern $\mathbf{V}_u^{\text{Pre}}$, social infrastructure reliance $\mathbf{SIR}_u$, and spatial context $\mathbf{SC}_u$, the task is to predict the post-event movement pattern $\mathbf{V}_u^{\text{Post}}$. Formally, we aim to learn a predictive function $f_\theta$ such that
\begin{equation}
    f_\theta: \left(\mathbf{V}_u^{\text{Pre}}, \mathbf{SIR}_u, \mathbf{SC}_u\right) \mapsto \hat{\mathbf{V}}_u^{\text{Post}},
\end{equation}
with the objective of minimizing the discrepancy between $\hat{\mathbf{V}}_u^{\text{Post}}$ and the ground truth $\mathbf{V}_u^{\text{Post}}$.


\section{Approach}


Our model adopts a UNet-based encoder–decoder architecture augmented with both spatial context and social infrastructure reliance conditioning to capture personalized and context-aware movement transitions. The spatial conditioning is performed by tiling the reliance vector $\mathbf{SIR}_u$ to match the spatial dimensions of the POI distribution map $\mathbf{SC}_u$. These two components are concatenated along the channel dimension to form the spatial condition map:
\begin{equation}
    \mathbf{C}_u^{\text{spatial}} = \text{Concat}(\mathbf{SC}_u, \mathbf{SIR}_u^{\text{tile}}),
\end{equation}
where $\mathbf{SIR}_u^{\text{tile}}$ is a spatially broadcasted version of $\mathbf{SIR}_u$. This condition map is processed alongside the binary pre-event movement pattern $\mathbf{V}_u^{\text{Pre}}$ through convolutional encoders, and the resulting features are fused and passed into the UNet backbone.
\begin{equation}
    \mathbf{F}_0 = \operatorname{Concat}\!\big(\operatorname{Conv}(\mathbf{C}_u^{\text{spatial}}),\ \operatorname{Conv}(\mathbf{V}_u^{\text{Pre}})\big).
\end{equation}
where $\mathbf{F}_0$ is the input of the first layer in cUNet. In parallel, social infrastructure reliance conditioning is performed by passing $\mathbf{SIR}_u$ through a multi-layer perceptron to generate layer-specific affine modulation parameters, where $l$ denotes the layer number:
\begin{equation}
    \boldsymbol{\gamma}_l, \boldsymbol{\beta}_l = \text{Proj}_l(\text{MLP}(\mathbf{SIR}_u)) \in \mathbb{R}^{C_l},
\end{equation}
which modulate each feature map $\mathbf{F}_l$ in the network via
\begin{equation}
    \tilde{\mathbf{F}}_l = \mathbf{F}_l \cdot (1 + \boldsymbol{\gamma}_l) + \boldsymbol{\beta}_l.
\end{equation}

The encoder path consists of four down-sampling blocks, each comprising convolution, batch normalization, ReLU activation, cross-attention, and feature modulation. A bottleneck block compresses the features, which are then passed through a symmetric decoder with transposed convolutions and skip connections. The final prediction $\hat{\mathbf{V}}_u^{\text{Post}}$ is obtained by applying a $1 \times 1$ convolution followed by a sigmoid activation applied to the decoder output.

We employ a weighted Binary Cross Entropy (BCE) loss to account for the imbalance between visited and unvisited grid cells. \textbf{Visited} cells are those visited before and after the event, while \textbf{unvisited} cells are those newly visited after the event. For each sample, the weight is dynamically adjusted based on the proportion of visits in the ground truth map:
\begin{equation}
    w_u = 1 + \left(w_{\text{max}} - 1\right)(1 - r_u),
\end{equation}
where $r_u$ is the ratio of visited cells and $w_{\text{max}}$ is a hyperparameter.
The training loss is computed as the mean of the BCE loss weighted by $w_u$ across the batch. This weighting scheme allows the model to focus more on capturing sparse, personalized movement behaviors. 
The model is optimized using Adam with learning rate decay via a plateau scheduler. To prevent gradient explosion, we apply norm-based gradient clipping at each training step.

\section{Evaluation}
\label{sec:evaluation}

\textbf{Experimental Setup.}
We evaluate the proposed method for both predictive performance and interpretability through four analyses: an ablation study on conditioning components, a spatial comparison of predicted versus ground-truth patterns, a case study of individuals with similar pre-event patterns but different SIR, and a sensitivity analysis of the spatial cropping parameter. 
Evaluation metrics include \textbf{Overall Accuracy}, \textbf{Visited Accuracy}, and \textbf{Unvisited Accuracy}, defined as follows: 
\begin{equation}
\text{Overall Accuracy} = \frac{\left|\hat{\mathbf{V}}_u^{\text{Post}} \cap \mathbf{V}_u^{\text{Post}}\right|}{\left|\mathbf{V}_u^{\text{Post}}\right|}
\end{equation}
\begin{equation}
\text{Visited Accuracy} = \frac{\left|\hat{\mathbf{V}}_u^{\text{Post}} \cap \mathbf{V}_u^{\text{Post}} \cap \mathbf{V}_u^{\text{Pre}}\right|}{\left|\mathbf{V}_u^{\text{Post}} \cap \mathbf{V}_u^{\text{Pre}}\right|}
\end{equation}
\begin{equation}
\text{Unvisited Accuracy} = \frac{\left|\hat{\mathbf{V}}_u^{\text{Post}} \cap \mathbf{V}_u^{\text{Post}} \cap \left(\mathbf{V}_u^{\text{Pre}}\right)^c\right|}{\left|\mathbf{V}_u^{\text{Post}} \cap \left(\mathbf{V}_u^{\text{Pre}}\right)^c\right|}
\end{equation}
where $\left(\mathbf{V}_u^{\text{Pre}}\right)^c$ denotes the cells not visited before the event.

\textbf{Dataset Description.}  
We use the YJMob100K dataset \cite{yabe_yjmob100k_2024}, which records the movements of 25,000 anonymized individuals in Japan at 30-minute intervals over 75 days (60 pre-event and 15 post-event). Each record is mapped to a $200 \times 200$ spatial grid with 85 categories of POIs. For this study, we sample 20,000/1,000/1,000 individuals for training/validation/testing, and define a $100 \times 100$ grid $\mathcal{M}_u$ for each individual centered on their most frequent day/night locations during the pre-event period. This provides individual-specific pre- and post-event movement patterns $\mathbf{V}_u^{\text{Pre}}, \mathbf{V}_u^{\text{Post}}$, spatial context $\mathbf{SC}_u \in \mathbb{R}^{K \times 100 \times 100}$, and an SIR vector $\mathbf{SIR}_u \in \mathbb{R}^K$, which are used as model inputs.


\textbf{Effects of Conditioning Components.}  
To assess the contributions of spatial and SIR conditioning, we implement several model variants incrementally. The \textbf{Baseline (Plain UNet)} uses only the pre-event movement pattern $\mathbf{V}_u^{\text{Pre}}$ as input, without conditioning on $\mathbf{SC}_u$ or $\mathbf{SIR}_u$. Adding spatial context yields \textbf{+Spatial}, which incorporates the POI distribution map $\mathbf{SC}_u$ through spatial concatenation and convolutional encoding. Extending this, \textbf{+Spatial+SIR (Concat)} concatenates the SIR vector with $\mathbf{SC}_u$, while \textbf{+Spatial+SIR (Modulation)} instead applies feature modulation with $\mathbf{SIR}_u$ at each UNet block. Finally, the \textbf{Full Model} integrates both reliance mechanisms—concatenation and modulation—with spatial context.
As shown in Table~\ref{tab:ablation}, incorporating spatial and behavioral conditioning consistently improves performance. Notably, SIR conditioning enhances accuracy in unvisited areas, while the \textbf{Full Model} achieves the highest overall accuracy but not the best performance for visited cells, suggesting that SIR promotes more conservative yet precise predictions of movement shifts under disruption.

\begin{table}[ht]
\centering
\caption{Ablation results on individual movement pattern \textbf{shift} prediction}
\label{tab:ablation}
\begin{tabular}{>{\raggedright\arraybackslash}p{3.5cm}ccc}
\toprule
\textbf{Model Variant} & \textbf{Overall} 
& \textbf{Visited} 
& \textbf{Unvisited} \\
\midrule
Baseline (Plain UNet) 
& 0.9722 & 0.7537 & 0.9732 \\
+Spatial 
& 0.9708 & \textbf{0.7848} & 0.9717 \\
+Spatial+SIR (Concat) 
& 0.9738 & 0.7812 & 0.9747 \\
+Spatial+SIR (Modulation) 
& 0.9745 & 0.7828 & 0.9753 \\
Full Model 
& \textbf{0.9752} & 0.7779 & \textbf{0.9761} \\
\bottomrule
\end{tabular}
\end{table}

\begin{figure}[ht]
    \centering    
    \includegraphics[width=0.45\textwidth]{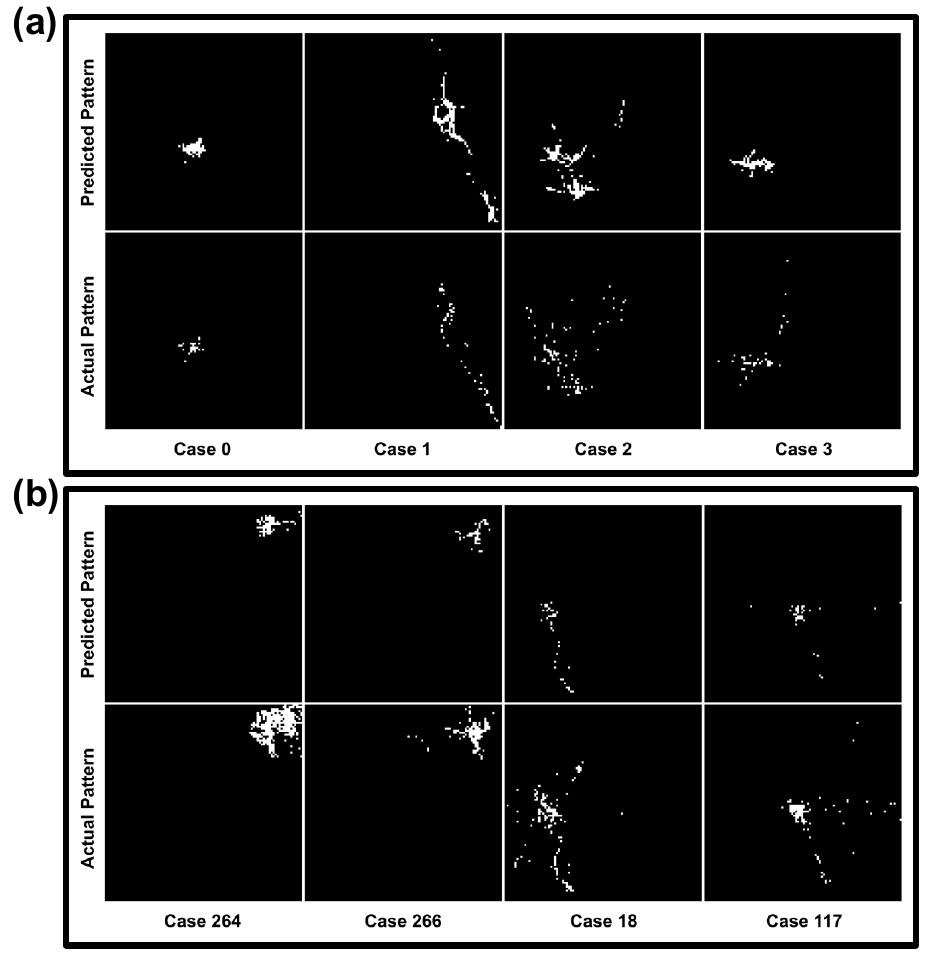}
    \caption{Predicted post-event movements of (a) sample individuals, and (b) individuals with similar pre-event patterns but different SIR}
    \label{fig:vis_examples}
\end{figure}

\textbf{Spatial Alignment with Observed Movement Patterns.}
We evaluate the model’s ability to capture post-event spatial patterns by comparing predicted and ground-truth movements.  Figure~\ref{fig:vis_examples}a shows results for four randomly selected individuals, where the predicted patterns closely align with the actual ones, demonstrating the model’s ability to replicate complex spatial transitions. However, limitations remain in modeling infrequent visitations. For instance, in Case 3, our model fails to predict isolated new visits in the central and upper regions of the grid. This shortfall likely reflects the irregular nature of such visits, which introduce noise that spatial context and SIR conditioning may not fully capture.

\textbf{Differentiating Movement Patterns for Individuals with Similar Pre-Event Patterns but Distinct SIR.}
To test whether our model can reflect behavioral divergence conditioned on SIR, we identify pairs of individuals with high similarity in pre-event movement patterns (cosine similarity > 0.3) but low similarity in SIR. Despite scores of 0.3104 (cases 264/266) and 0.3201 (cases 18/117), the model produces distinct post-event patterns (Figure~\ref{fig:vis_examples}b): individuals with stronger reliance on certain POIs (e.g., health services or retail) form new spatial clusters, while their counterparts do not.


\textbf{Sensitivity to Cropping Size for Spatial Context Around Home Location.}
We test spatial context sizes of $50 \times 50$, $100 \times 100$, and $150 \times 150$. The default $100 \times 100$ achieves higher accuracy than $50 \times 50$ (Table~\ref{tab:crop_sensitivity}), while enlarging to $150 \times 150$ yields no significant further gains. This suggests $100 \times 100$ offers the best trade-off between contextual coverage and focus on local areas.

\begin{table}[ht]
\centering
\caption{Sensitivity to crop size of spatial context.}
\label{tab:crop_sensitivity}
\begin{tabular}{lccc}
\toprule
\textbf{Model} & \textbf{Overall} 
& \textbf{Visited} 
& \textbf{Unvisited} \\
\midrule
$50 \times 50$ & 0.9223 & 0.6112 & 0.9271 \\
$100 \times 100$ & 0.9752 & 0.7779 & 0.9761 \\
$150 \times 150$ & 0.9783 & 0.7260 & 0.9776 \\
\bottomrule
\end{tabular}
\end{table}

\section{Conclusion and Future Works}

We present an approach for predicting the shifts of individual movement patterns under urban disruptions by incorporating two critical conditioning factors: social infrastructure resilience (SIR) and local spatial context. We employ a cUNet model to effectively capture heterogeneous behavioral responses across individuals by learning from their pre-event movement patterns and environmental context. Our evaluation on a large-scale real-world mobility dataset demonstrates that SIR and spatial context improve the model’s ability to predict nuanced changes in post-disruption behaviors. By complementing population-level analysis, our framework contributes to a more behavior-aware and context-adaptive understanding of urban resilience, enabling the identification of vulnerable individuals and informing targeted interventions.

Future work will extend this framework in several directions. First, we aim to incorporate temporal dynamics by modeling how individuals adapt their movement over time following a disruption, rather than relying on static pre- and post-event snapshots. Second, our framework can be enhanced with more advanced conditional generative models (e.g., denoising diffusion models) to sample mobility guided by context. Finally, incorporating uncertainty estimation and causal inference could make the model more robust and actionable for real-world decision-making in urban planning and disaster response.


\bibliographystyle{ACM-Reference-Format}
{\bibliography{SIGSPATIAL2025}}

\end{document}